\documentclass[letterpaper, 10 pt, conference]{ieeeconf}  

\IEEEoverridecommandlockouts                              

\usepackage{caption}
\usepackage{graphicx}
\usepackage{stfloats}
\usepackage{float} 
\usepackage{subcaption}
\usepackage{color}
\usepackage{amsmath}
\usepackage{cite}
\usepackage{makecell}
\usepackage{booktabs}
\usepackage{multirow}
\usepackage{diagbox}
\makeatletter
\let\NAT@parse\undefined
\makeatother
\usepackage{hyperref}
\hypersetup{pdfstartview=FitH,
            colorlinks=true,
            linkcolor=red,
            anchorcolor=blue,
            citecolor=green
            }

\begin{document}
\title{\LARGE \bf
Continuity-Aware Latent Interframe Information Mining\\ for Reliable UAV Tracking
}
\author{Changhong Fu$^{1,}$*, Mutian Cai$^{1}$, Sihang Li$^{1}$, Kunhan Lu$^{1}$, Haobo Zuo$^{1}$, and Chongjun Liu$^{2}$
\thanks{*Corresponding author}
\thanks{$^{1}$C. Fu, M. Cai, S. Li, K. Lu, and H. Zuo are with the School of Mechanical Engineering, Tongji University, Shanghai 201804, China. {\tt\small changhongfu@tongji.edu.cn}}%
\thanks{$^{2}$C. Liu is with the College of Mechanical and Electrical Engineering, Harbin Engineering University, Harbin 150001, China. {\tt\small chongjunliu@hrbeu.edu.cn}}%
}


\maketitle
\thispagestyle{empty}
\pagestyle{empty}

\begin{abstract}
Unmanned aerial vehicle (UAV) tracking is crucial for autonomous navigation and has broad applications in robotic automation fields.
However, reliable UAV tracking remains a challenging task due to various difficulties like frequent occlusion and aspect ratio change.
Additionally, most of the existing work mainly focuses on explicit information to improve tracking performance, ignoring potential interframe connections.
To address the above issues, this work proposes a novel framework with continuity-aware latent interframe information mining for reliable UAV tracking, \textit{i.e.}, ClimRT.
Specifically, a new efficient continuity-aware latent interframe information mining network (ClimNet) is proposed for UAV tracking, which can generate highly-effective latent frame between two adjacent frames. 
Besides, a novel location-continuity Transformer (LCT) is designed to fully explore continuity-aware spatial-temporal information, thereby markedly enhancing UAV tracking.
Extensive qualitative and quantitative experiments on three authoritative aerial benchmarks strongly validate the robustness and reliability of ClimRT in UAV tracking performance. 
Furthermore, real-world tests on the aerial platform validate its practicability and effectiveness.
The code and demo materials are released at \href{https://github.com/vision4robotics/ClimRT}{https://github.com/vision4robotics/ClimRT}.

\end{abstract}

\section{Introduction}

Unmanned aerial vehicle (UAV) tracking has drawn considerable attention due to its prosperous applications in robotic automation, \textit{e.g.}, automatic patrol~\cite{SY2021autopatrol}, robot cinematography~\cite{BA2021roboph}, and aerial manipulator~\cite{OZ2021FLYHAND}.
Despite remarkable advancements, efficient and accurate UAV tracking remains a challenging task due to several inherent difficulties.
Due to the limited aerial perspective, reliable object information is frequently unavailable in complex aerial scenarios.

In the visual tracking community, Siamese trackers~\cite{LB2016SiamFC, Li2019TADT, Li2019SiamRPN++} have achieved impressive performance account of superior feature extraction ability. 
However, the object information that is obtained solely from the template and search frames can be limited, especially in complex aerial scenarios, such as occlusion and aspect ratio change, \textit{etc}.
To improve the robustness of tracking, more reliable additional information needs to be explored.
Despite many works introducing explicit temporal information~\cite{Zhang2019UpdateNet,Yan2021LST}, they still ignore mining latent robust interframe information.
Furthermore, even if more feature information is obtained, convolutional feature maps still suffer from local sensitivity and translation invariance.
Since convolutional neural network (CNN) cannot extract feature connections between global information, it is challenging to utilize all available information efficiently.
Recently, Transformer~\cite{VA2017Attention} has shown great potential in many fields with its attention mechanism. 
Transformer enables the network to effectively execute global information integration, increasing the robustness of object tracking.
Nonetheless, when the UAV is operating at elevated velocities, Transformer-based trackers still suffer from discontinuous or lost object information due to issues like aspect ratio change and occlusion.


\begin{figure}[tbp] 
 \centering
 
 \includegraphics[width=0.93\linewidth]{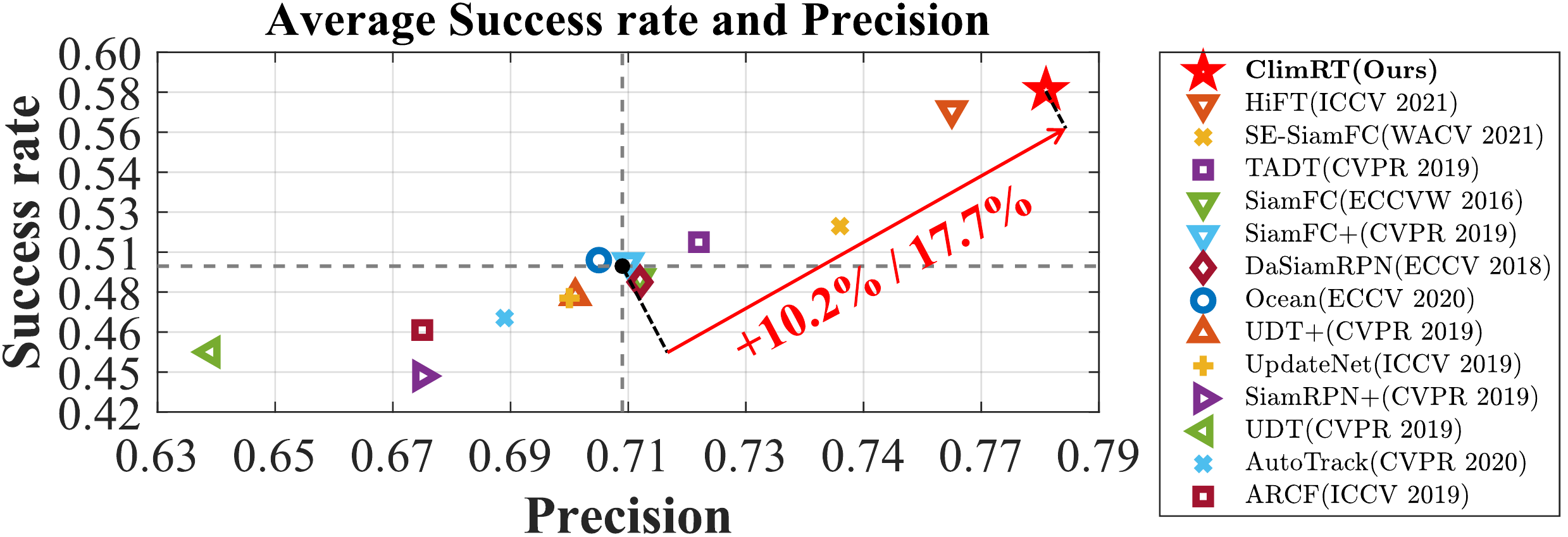}
 \setlength{\abovecaptionskip}{7pt}
 \caption
 {
  Overall comparison of average success rate and precision between ClimRT and other 13 state-of-the-art (SOTA) trackers on three well-known benchmarks~\cite{MM2016UAV123,Fu2022UAVTrack112}.
  ClimRT gains first place in both precision and success rate, surpassing the average performance of 13 trackers (\textbf{black dot}) by \textbf{10.2\%} and \textbf{17.7\%} respectively.
 }
 \label{1}
 \vspace{-23pt}
\end{figure}

For reliable UAV tracking, this work aims to mine the interframe information to introduce sufficient spatial-temporal information.
Through the mining of interframe information, temporal information that exists between frames can be fully explored and utilized.
Additionally, introducing interframe information can efficiently enhance information continuity.
When encountering a common object shape deformation or occlusion, the introduction of interframe information can smooth the mutation process, thus extracting more reliable information and enhancing tracking robustness.

\begin{figure*}[htbp]
	\centering
	\includegraphics[width=0.92\linewidth]{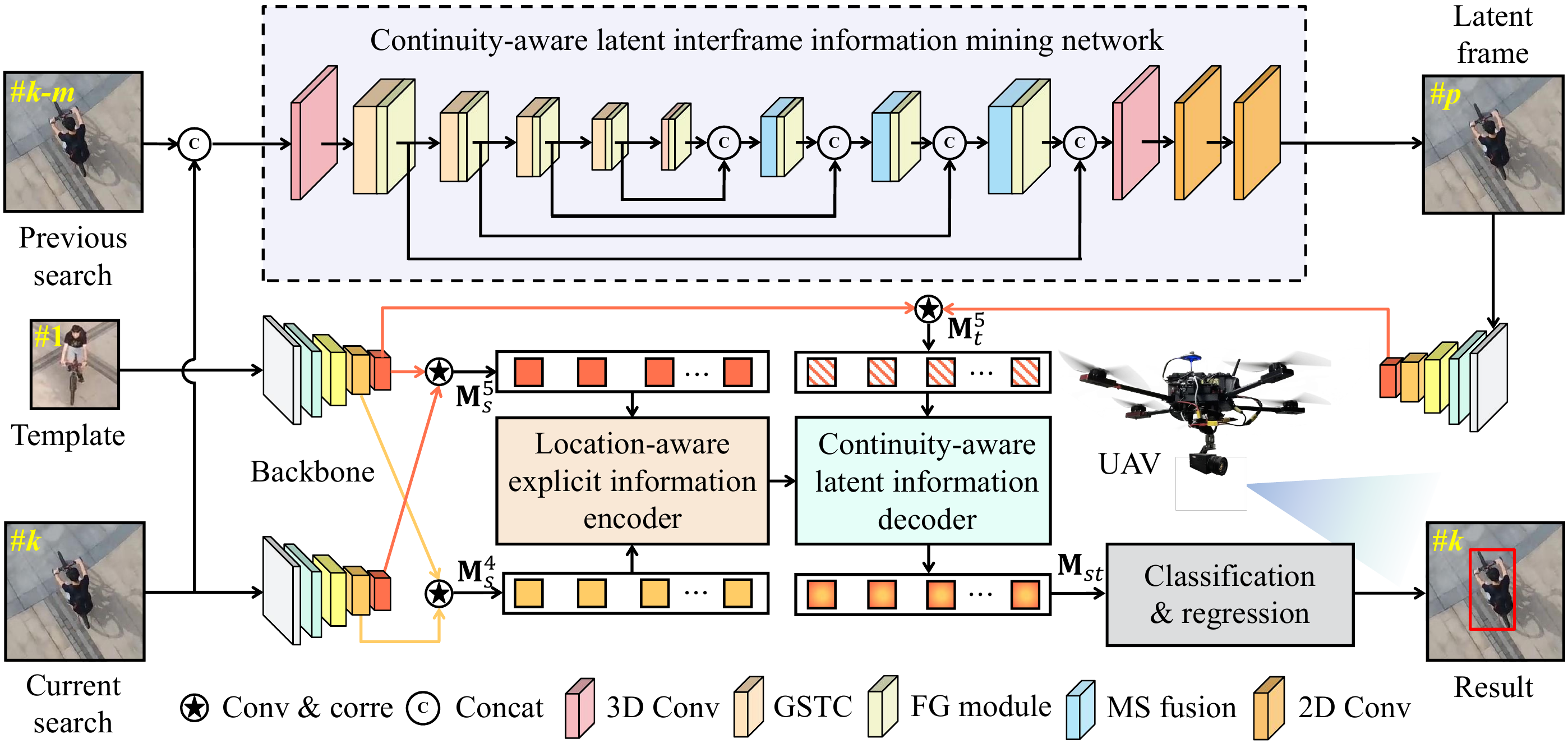}
	\caption{Overview of the proposed ClimRT tracker. Two arrows with different colors represent the workflow of features from different layers respectively. Best viewed in color. (Sequence courtesy of benchmark UAVTrack112~\cite{Fu2022UAVTrack112}.)}
	\label{2}
	\vspace{-18pt}
\end{figure*}

Interframe information mining, such as motion prediction~\cite{Wang2020MotionPI}, optical flow prediction~\cite{Xing2021Real-timeOT}, has been utilized to extract explicit information.  
However, the full potential of latent interframe information in UAV tracking remains unexplored.
Thus, this work not only introduces explicit information but also attempts to explore the latent interframe information.
Specifically, a novel continuity-aware latent interframe information mining network (ClimNet) is proposed to provide additional reliable information while improving information continuity and effectively solving the problem of sudden appearance changes. 
Additionally, a location-continuity Transformer (LCT) is designed for effective feature fusion and global relationship modeling.
Finally, reliable information is introduced for robust tracking.
Fig.~\ref{1} illustrates the satisfactory performance of ClimRT in UAV tracking compared to the state-of-the-art (SOTA) trackers.
The main contributions of this work are as follows:

\begin{itemize}

\item  An efficient ClimNet suitable for UAV tracking is proposed, which can provide sufficient and reliable object information to enhance robustness.

\item  Two Conv blocks are introduced to improve the feature extraction and fusion capabilities for continuity-aware latent interframe information.

\item  A novel LCT is designed for spatial-temporal information integration and continuity enhancement to improve the accuracy of UAV tracking.

\item  Comprehensive evaluations on three authoritative aerial benchmarks have validated the promising performance of ClimRT compared with other SOTA trackers. Real-world tests are conducted, demonstrating the superior effectiveness of ClimRT in real-world circumstances. 
\end{itemize}

\section{Related Works}

\vspace{-2pt}
\subsection{UAV Object Tracking}
\vspace{-3pt}
In recent years, Siamese object tracking algorithms are becoming increasingly popular~\cite{LB2016SiamFC,Zhu2018DaSiamRPN,fu2022siamese,li2022pvt++}. 
These methods generally consist of two steps: a backbone network to extract image features, and a correlation-based network to compute the similarity between template and search features. 
However, it is simple for the CNN feature map to be deficient in global information, which restricts the tracking algorithm's efficacy.
Using a larger kernel size~\cite{Li2019SiamRPN++} or improving convolution method~\cite{yu2015kdjj} may overcome this shortcoming to some extent.
Nevertheless, it is difficult to ensure the reliability of tracking using only local feature information.
Differently, this work proposes a novel location-continuity Transformer for efficient multi-level feature fusion and global information integration.
Consequently, the strengths of CNN and Transformer can be combined for robust UAV tracking.

\vspace{-2pt}
\subsection{ViT in UAV Object Tracking}
\vspace{-3pt}
Transformer~\cite{VA2017Attention} was firstly proposed for machine translation based on the attention mechanism. 
Due to its good performance in global information integration, Transformer has been extended to various computer vision fields~\cite{Dosovitskiy2021AnII,Xie2021Segformer,Yang2020LTT}.
TransT~\cite{Chen2021TransT} designs two attention modules for object tracking, achieving impressive performance.
HiFT~\cite{Cao2021HiFT} modifies the structure of Transformer for efficient feature fusion, which further enhances tracking robustness.
TransTrack~\cite{sun2020transtrack} introduces Transformer to multi-object tracking and achieves promising performance.
However, many Transformer-based tracking methods still have the limitation of using only the information from the original template.
Limited object information may lead to the degradation of UAV tracking performance in complex scenarios. 
To further improve the robustness and reliability of UAV tracking, this work supplements sufficient reliable information through continuity-aware latent interframe information mining.


\vspace{-2pt}
\subsection{Spatial-Temporal Information Introduction}
\vspace{-3pt}
Most of the existing Siamese trackers~\cite{LB2016SiamFC,Li2018SiamRPN,Li2019SiamRPN++,Zhang2020ocean} regard object tracking as a template-matching process between template and search. 
However, during the tracking process, challenges such as occlusion and aspect ratio change can lead to alterations in the appearance of the object.
Under these circumstances, depending solely on the limited object information extracted from the template may significantly deteriorate the tracking performance.
Several researches~\cite{Zhang2019UpdateNet,wang2021trsiam,cao2022tctrack} have also noticed that and attempted to introduce additional information in the tracking process to increase robustness. 
UpdateNet~\cite{Zhang2019UpdateNet} proposes a template dynamic update method to extract more object temporal information during tracking.
TrSiam~\cite{wang2021trsiam} uses template patches to explore the temporal relationship of multiple frames.
Nevertheless, these works only explore explicit information and ignore robust latent information between frames.
Differently, this work proposes a novel continuity-aware latent interframe information mining network (ClimNet). 
By exploring the latent information contained between adjacent frames, ClimNet can provide sufficient and reliable object information to make UAV tracking reliable and robust.

\section{Proposed Method}

The workflow of ClimRT is shown in Fig.~\ref{2}. 
It mainly consists of two novel parts, \textit{i.e.}, ClimNet and LCT. 
\vspace{-15pt}

\subsection{Continuity-Aware Latent Interframe
Information Mining Network}
\begin{figure}[tbp]
	\centering
	\includegraphics[width=0.75\linewidth]{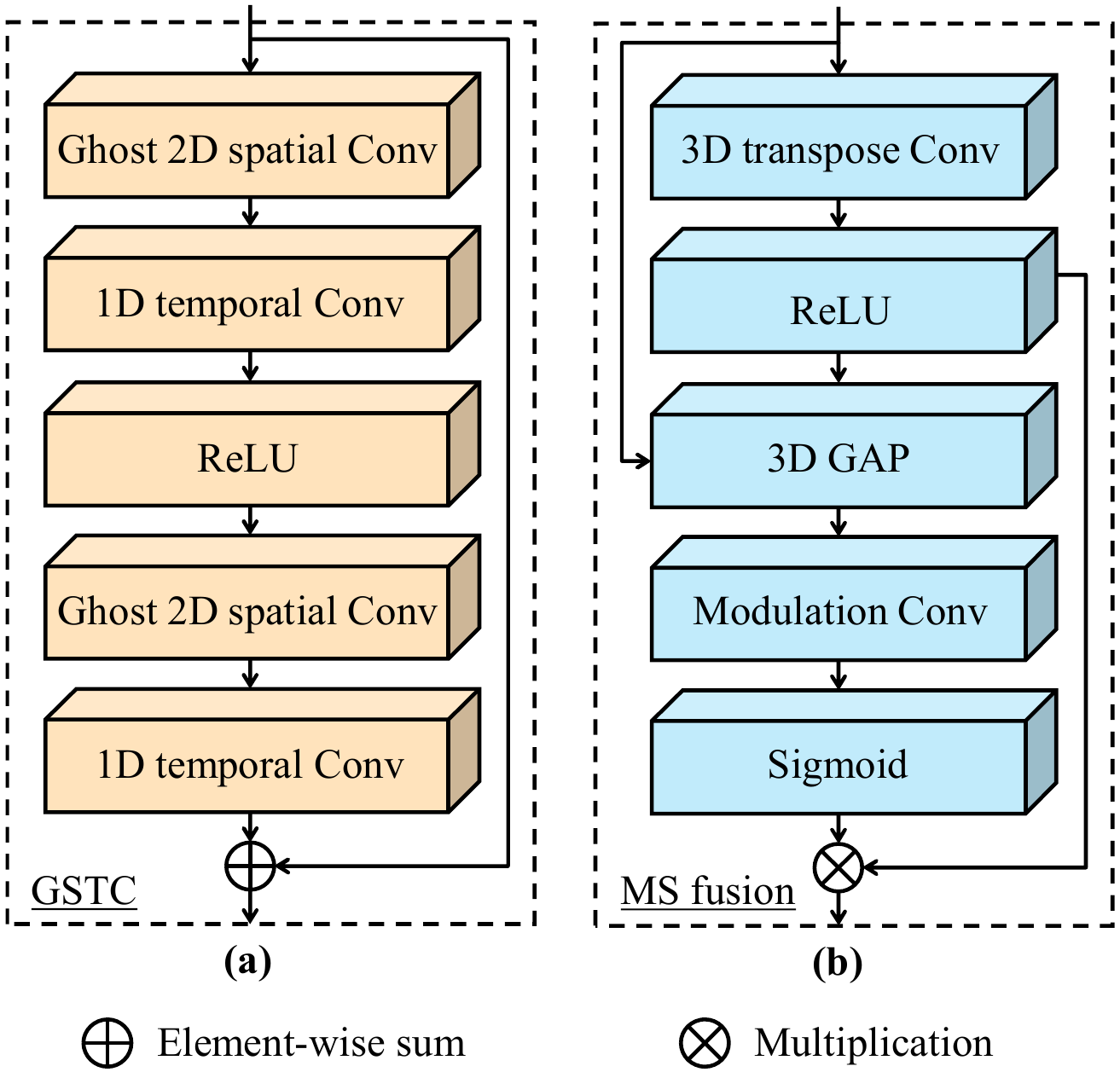}
	\caption{Structure of the two novel blocks. The left sub-window shows the components of the GSTC block while the right one illustrates the MS fusion block.}
	\label{3}
	\vspace{-18pt} 
\end{figure}

\vspace{-3pt}
An existing UNet-based interframe information mining method~\cite{kalluri2020flavr} is conducive to exploring interframe information.
Inspired by it, ClimNet is proposed to explore connections between frames to provide sufficient and reliable information.
Two novel blocks are introduced for more efficient information mining and restoration of object features.
Note that the last three ghost spatial-temporal Conv (GSTC) blocks perform the downsampling operation while the multi-scale (MS) fusion blocks perform the upsampling operation. 
The detailed structures of the two novel blocks are presented in Fig.~\ref{3}. 
In order to fully mine and fuse multi-scale interframe information, skip connections are employed between the information of the same scale.
Besides, feature gating modules~\cite{Xie2018S3D} are added after each intermediate block to learn more robust features.

\subsubsection{Ghost Spatial-Temporal Conv}
Each GSTC block consists of two Ghost 2D spatial Conv layers, two 1D temporal Conv layers, and a ReLU function.
3DConv enhances the ability in modeling temporal abstractions while introducing a large amount of computation.
Thus, this work combines GhostNet~\cite{Han2020GhostNet} with S3D~\cite{Xie2018S3D} for an efficient feature extraction module.
The main process of the Ghost 2D spatial Conv layer is to obtain the feature map with fewer channels to prevent feature redundancy and then generate the final desired feature map through cheap mapping.
Temporal information is then integrated through the 1D temporal Conv layer, which can be formulated as:
\begin{equation}
\begin{aligned}
{\rm \bf M}_z = &~{\rm G2SC}({\rm \bf M}_i) ={\rm map}({\rm 3DSConv}({\rm \bf M}_i))\quad,\\
{\rm \bf M}_o = &~{\rm 1DTConv}({\rm \bf M}_z))\quad,
\end{aligned}
\end{equation}
where $\rm map$ denotes the cheap mapping operation implemented by a convolutional layer, ${\rm \bf M}_i$ and ${\rm \bf M}_o$ represent the input and output of two neighboring Conv modules respectively.
${\rm G2SC}$ and ${\rm 1DTConv}$ correspond to Ghost 2D spatial Conv layer and 1D temporal Conv layer.
Note that $\rm 3DSConv$ and $\rm 1DTConv$ are implemented by two 3D convolutional layers with the kernel size of $1\times3\times3$ and $3\times1\times1$ respectively.
For convenience, $\rm STConv$ represents a single 2+1D convolution, then the workflow of the entire GSTC block can be expressed as:
\begin{equation}
{\rm \bf F}_o = {\rm \bf F}_i+{\rm STConv}({\rm ReLU}({\rm STConv}({\rm \bf F}_i)))\quad,
\end{equation}
where ${\rm \bf F}_i$ and ${\rm \bf F}_o$ represent the input and output of GSTC block respectively.
Note that the downsampling operation can be implemented by adjusting kernel size and stride.

\noindent \textbf{\textit{Remark 1:}}
Attributing to ghost mapping and spatial-temporal split convolution, GSTC achieves robust feature extraction capability, obtaining promising tracking performance with the calculation suitable for UAV tracking.

\subsubsection{Multi-Scale Fusion}
In MS fusion block, 3D transpose Conv (${\rm 3DTransConv}$) is employed for upsampling.
Besides, a novel residual connection is designed to enhance the ability to restore the detailed characteristics of the object.
Specifically, the small-scale information is first input to global average pooling ($\rm GAP$) and Modulation Conv layer $(\rm ModulConv)$ while upsampling.
Thus, the Modulation Conv layer able to further fuse the small-scale feature as a coefficient with the upsampling result.
This coefficient ${\rm \bf W}_s$ can be obtained as follows:
\begin{equation}
{\rm \bf W}_s = {\rm Sigmoid}({\rm ModulConv}({\rm GAP}({\rm \bf F}_i^{'})))\quad.
\end{equation}

After average pooling and channel modulation, the information from the small scale can effectively enrich the feature information of the object.
Generally, the workflow of the entire MS fusion block can be expressed as:
\begin{equation}
{\rm \bf F}_o^{'} = {\rm \bf W}_s * {\rm ReLU}({\rm 3DTransConv}({\rm \bf F}_i^{'}))\quad,
\end{equation}
where ${\rm \bf F}_i^{'}$ and ${\rm \bf F}_o^{'}$ represent the input and output feature maps of the MS fusion block respectively.

\noindent \textbf{\textit{Remark 2:}}
Multi-scale fusion blocks can fully integrate features of different scales.
Subsequently, more multi-scale feature information can be mined and retained to do better in detail restoration.
Consequently, the UAV tracking network can extract more object details and improve accuracy.

\subsubsection{Feature Transform Layers}
Two 2D convolutional layers are employed to transform the last output feature map from 3D to 2D which is suitable for tracking.
The first 2D Conv layer can fuse temporal information along the time dimension. 
Subsequently, the second 2D Conv layer can integrate and restore information to obtain the predicted latent frame with size $H\times W\times 3$, where $H$ and $W$ are the spatial dimensions of the input frame.

\noindent \textbf{\textit{Remark 3:}}
Through ClimNet, sufficient interframe information is mined and integrated.
Therefore, reliable information is introduced to improve information continuity and effectively mitigate the negative impact of object shape change, strengthening the UAV tracking-oriented task robustness.

\vspace{-3pt}
\subsection{Object Tracking Network}
\vspace{-3pt}
\subsubsection{Feature Extraction Network}
In order to provide a clear explanation of the method, template, latent frame, and current search are denoted by $\bf Z$, $\bf L$, and $\bf X$, respectively. $\varphi_k(\bf Z)$ represents the $k$-th layer output of the search branch.

\subsubsection{Location-Continuity Transformer}
\begin{figure}[tbp]
	\centering
	\includegraphics[width=0.75\linewidth]{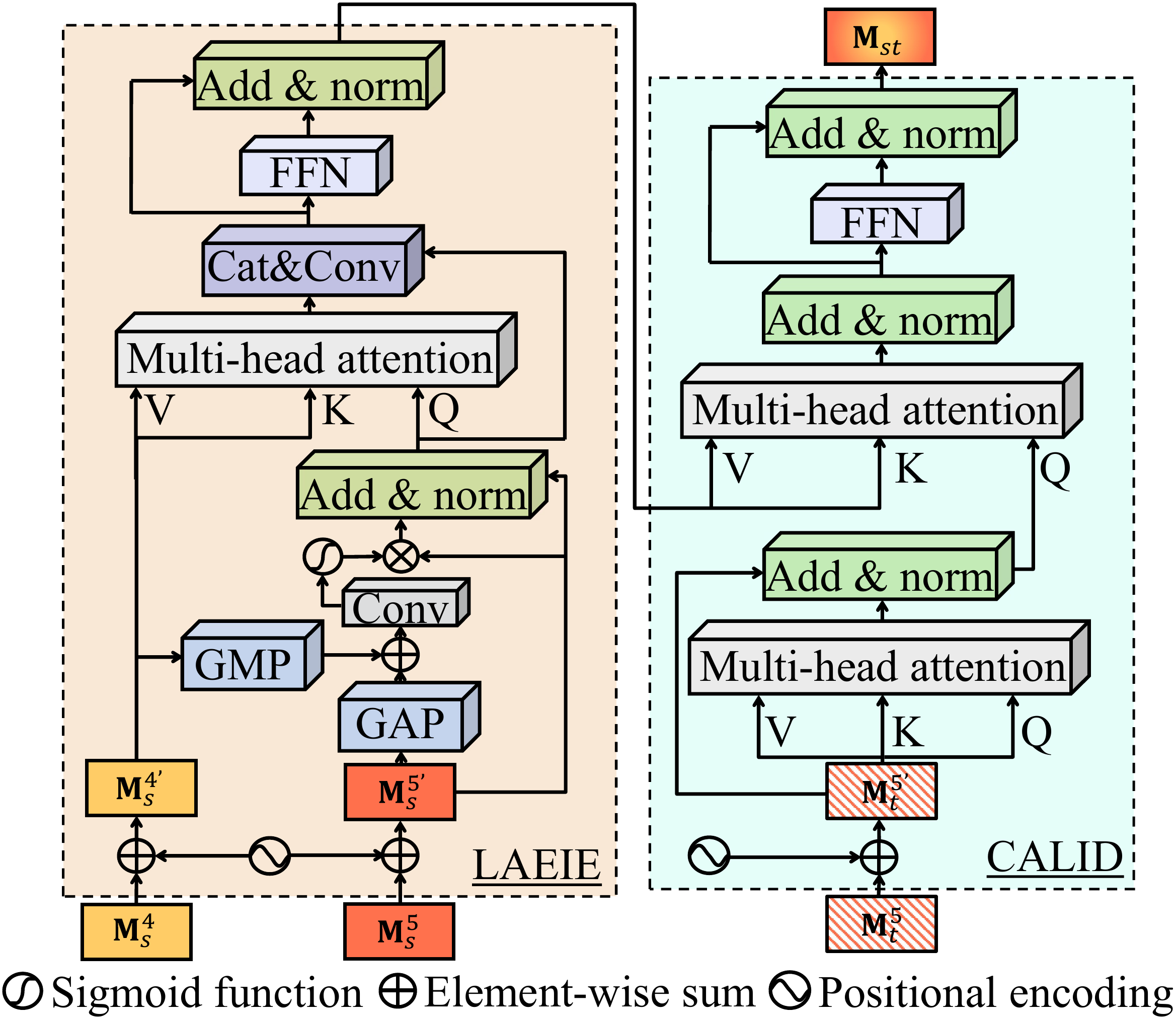}
	\caption{Detailed workflow of LCT. The left sub-window illustrates the location-aware explicit information encoder. The right one shows the structure of the continuity-aware latent information decoder. Best viewed in color.}
	\label{4}
    \vspace{-18pt} 
\end{figure}

The designed LCT can be divided into two parts: location-aware explicit information encoder (LAEIE) and continuity-aware latent information decoder (CALID).
Its structure is exhibited in Fig.~\ref{4}.
Before being fed into LCT, the feature map will be convoluted and reshaped to ${\rm \bf M}^i\in {\rm R}^{WH \times C}$. 
The formulas to obtain multi-layer similarity vector ${\rm \bf M}^i$ are as follows: ${\rm \bf M}_s^i = \varphi_i({\bf Z}) \star \varphi_i({\bf X}), i=4, 5$, ${\rm \bf M}_t^i = \varphi_i({\bf Z}) \star \varphi_i({\bf L}), i=5$, where $\star$ represents convolution and cross-correlation operation. 
Then, ${\rm \bf M}_s^{4'}$, ${\rm \bf M}_s^{5'}$ and ${\rm \bf M}_t^{5'}$ can be obtained by ${\rm \bf M}_s^4$, ${\rm \bf M}_s^4$ and ${\rm \bf M}_t^5$ plus a positional encoding.

\subsubsection*{\bf Location-Aware Explicit Information Encoder}
In order to fully fuse multi-layer features, the features of the fourth and fifth layers are preliminarily fused.
Global average pooling ($\rm GAP$) and global maximum pooling ($\rm GMP$) are adopted to keep information and obtain a coefficient after the convolution layer and sigmoid function:
\begin{equation}
\setlength{\abovedisplayskip}{3pt}
{\rm \bf W}^{'} = {\rm Sigmoid}({\rm Conv}({\rm GMP}({\rm \bf M}_s^{4'}) + {\rm GAP}({\rm \bf M}_s^{5'})))\quad.
\setlength{\belowdisplayskip}{3pt}
\end{equation}

Then, the fusion result of multi-layer features ${\rm \bf M}_s^{4\&5}$ is obtained by ${\rm \bf M}_s^{5'}$ and ${\rm \bf W}^{'}$:
\begin{equation}
\setlength{\abovedisplayskip}{3pt}
{\rm \bf M}_s^{4\&5} = {\rm \bf M}_s^{5'} + {\rm \bf W}^{'} * {\rm \bf M}_s^{5'}\quad.
\setlength{\belowdisplayskip}{3pt}
\end{equation}

Finally, further feature fusion can be conducted through multi-head attention~\cite{VA2017Attention}.
${\rm \bf M}_s^{4\&5}$ is the query (\textbf{Q}) for shallower layer features ${\rm \bf M}_s^{4'}$.
After that, some adjustments are made to the output of multi-head attention by the convolution layer and feed-forward network (FFN).

\noindent \textbf{\textit{Remark 4:}}
Attributing to LAEIE, the explicit spatial information contained in multi-layers is fully explored, obtaining a feature map with more effective information.


\subsubsection*{\bf Continuity-Aware Latent Information Decoder}
Two multi-head attention are employed to perform efficient feature fusion.
After self-attention enhancement, ${\rm \bf M}_t^{5'}$ is further fused with the output of the encoder as \textbf{Q}.
Eventually, the decoded information ${\rm \bf M}_{st}$ with rich spatial-temporal information can be calculated through FFN and normalization.

\noindent \textbf{\textit{Remark 5:}}
CALID effectively fuses explicit spatial and latent temporal information, smoothens the movement and deformation of objects, and improves information continuity.
Besides, object information is effectively integrated, thereby enhancing tracking robustness and accuracy.

\vspace{-3pt}
\subsection{Loss Function for Training}
\vspace{-3pt}
Pixel-level loss is used to train ClimNet while the combination of several loss functions is used for tracking net:
\begin{equation}
\setlength{\abovedisplayskip}{3pt}
L_{ifm} = \frac{1}{N} \sum_{i=1}^{N} ||F_p^i - F_t^i||_1\quad,
\setlength{\belowdisplayskip}{3pt}
\end{equation}
where $F_p^i$ and $F_t^i$ are the predicted and ground-truth frames of the $i$-th training clip, $N$ represents the batchsize.
\begin{equation}
\setlength{\abovedisplayskip}{3pt}
L_{track} = \lambda_{1}L_{cls1} + \lambda_{2}L_{cls2} + \lambda_{3}L_{loc}\quad,
 \setlength{\belowdisplayskip}{3pt}
\end{equation}
where $L_{cls1}$, $L_{cls2}$, and $L_{loc}$ represent the cross-entropy, binary cross-entropy, and IoU loss. 
$\lambda_{1}$, $\lambda_{2}$, and $\lambda_{3}$ are the weights used to control the entire loss function.

\noindent \textbf{\textit{Remark 6:}}
Two classification branches are applied for accuracy.
The combination of two branches with different focuses can achieve the expected classification effect.

\section{Experiments}

\subsection{Implementation Details}
\vspace{-3pt}
Before commencing overall network training, this work pretrains the ClimNet and the object tracking network respectively. 
The equipment used for training is a PC with an Intel \textit{i}9-9920X CPU, 32GB RAM, and two NVIDIA TITAN RTX GPUs.
Specifically, three consecutive frames from the Vimeo-90K dataset \cite{Xue2019Vimeo90K} are used for training ClimNet for 100 epochs with 16 batchsize for 5 days.  
For the object tracking network, this work uses image pairs extracted from COCO~\cite{Lin2014COCO}, ImageNet VID~\cite{OR2015ImageVID} and GOT-10K~\cite{Huang2021got10k} to pre-train it for 30 epochs with the batchsize of 128 over a period of 20 hours.
The learning rate is initialized as $5 \times 10^{-4}$ and decreased in the log space from $10^{-2}$ to $10^{-4}$. 
After pre-training, the two networks are integrated for end-to-end training with the same training strategy as the pre-trained tracking network for 40 epochs in 2 days.

\vspace{-2pt}
\subsection{Evaluation on Aerial Benchmarks}
\vspace{-3pt}
To assess tracking performance, the one-pass evaluation (OPE) metrics~\cite{MM2016UAV123} including precision and success rate are applied.
For overall evaluation, ClimRT is tested on three authoritative aerial tracking benchmarks and comprehensively compared with 13 SOTA trackers including HiFT~\cite{Cao2021HiFT}, DaSiamRPN~\cite{Zhu2018DaSiamRPN}, UDT~\cite{Wang2019UDT}, UDT+~\cite{Wang2019UDT}, TADT~\cite{Li2019TADT}, SiamFC~\cite{LB2016SiamFC}, UpdateNet~\cite{Zhang2019UpdateNet}, SE-SiamFC~\cite{IS2021SESiam}, Ocean~\cite{Zhang2020ocean}, SiamFC+~\cite{Zhang2019DWSiam}, SiamRPN+~\cite{Zhang2019DWSiam}, ARCF~\cite{Huang2019ARCF}, and AutoTrack~\cite{Li2020autotrack}. For fairness, all the Siamese trackers adopt the same backbone, \textit{i.e.}, AlexNet~\cite{AK2017AlexNet}, pre-trained on ImageNet~\cite{OR2015ImageVID}.

\begin{figure*}[!t] 
 \centering
 \subfloat{
 \includegraphics[width=0.26\linewidth]{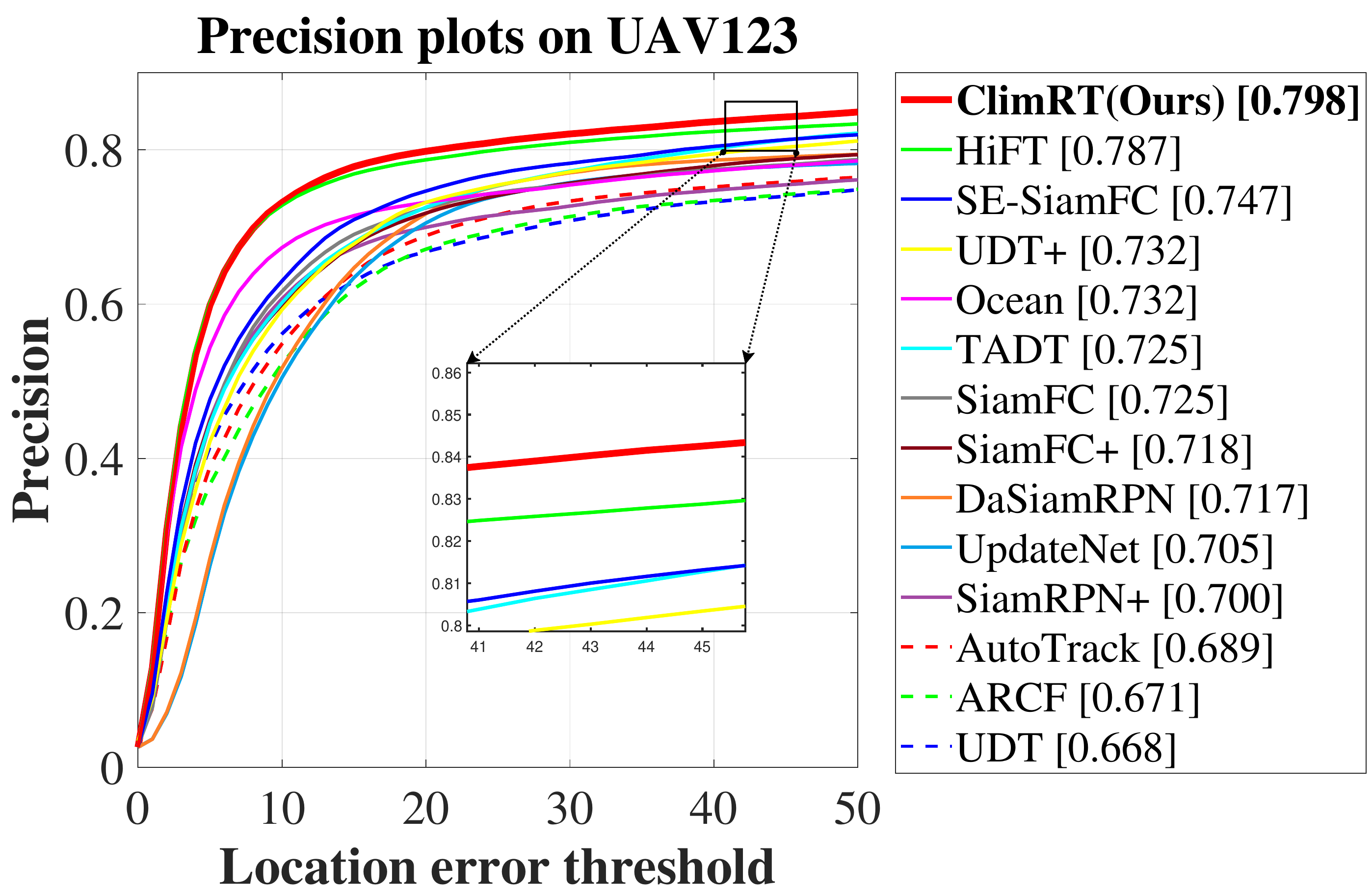}
 }
 \hfil
 \subfloat{
 \includegraphics[width=0.26\linewidth]{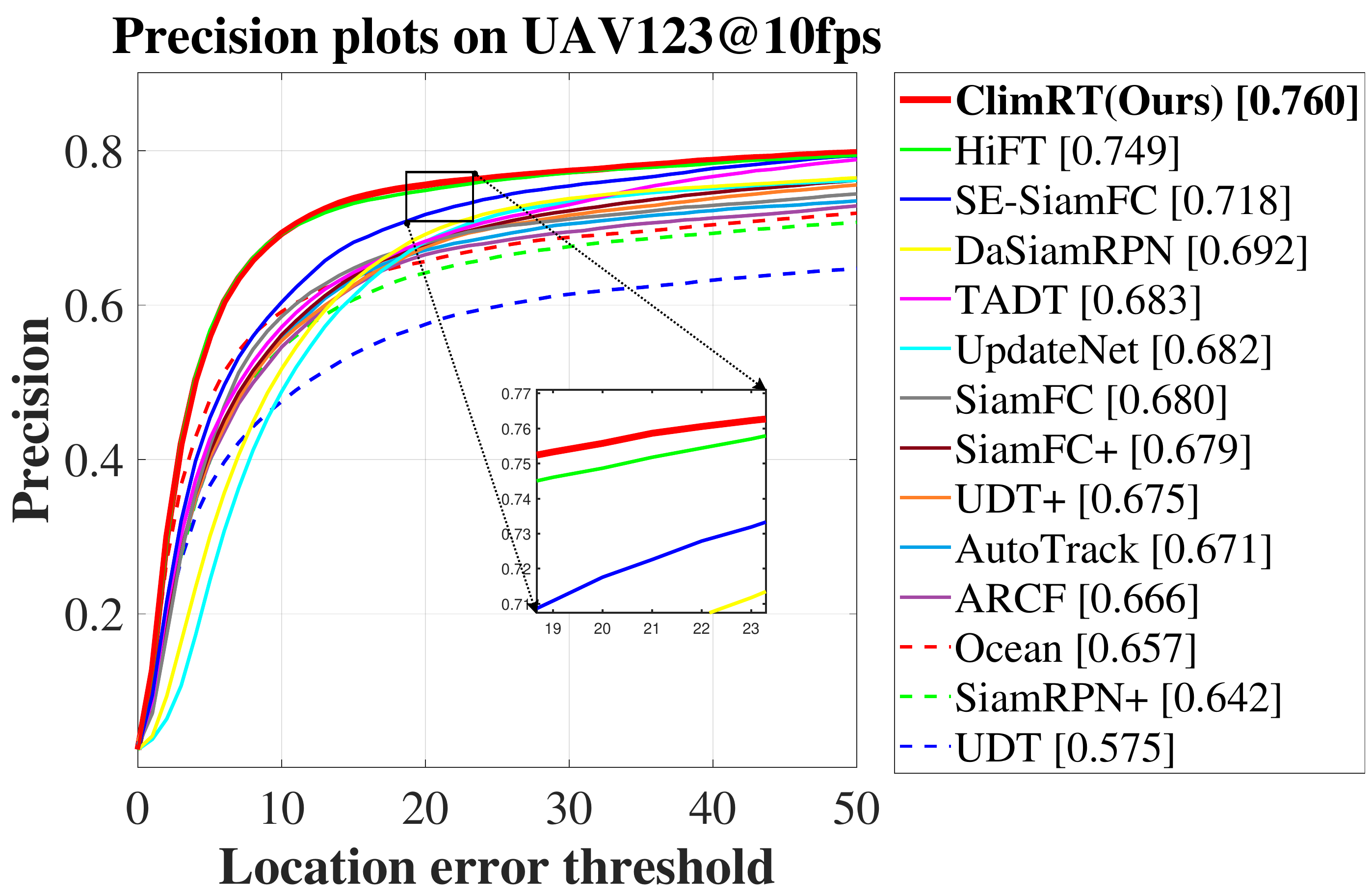}
 }
 \hfil
 \subfloat{
 \includegraphics[width=0.26\linewidth]{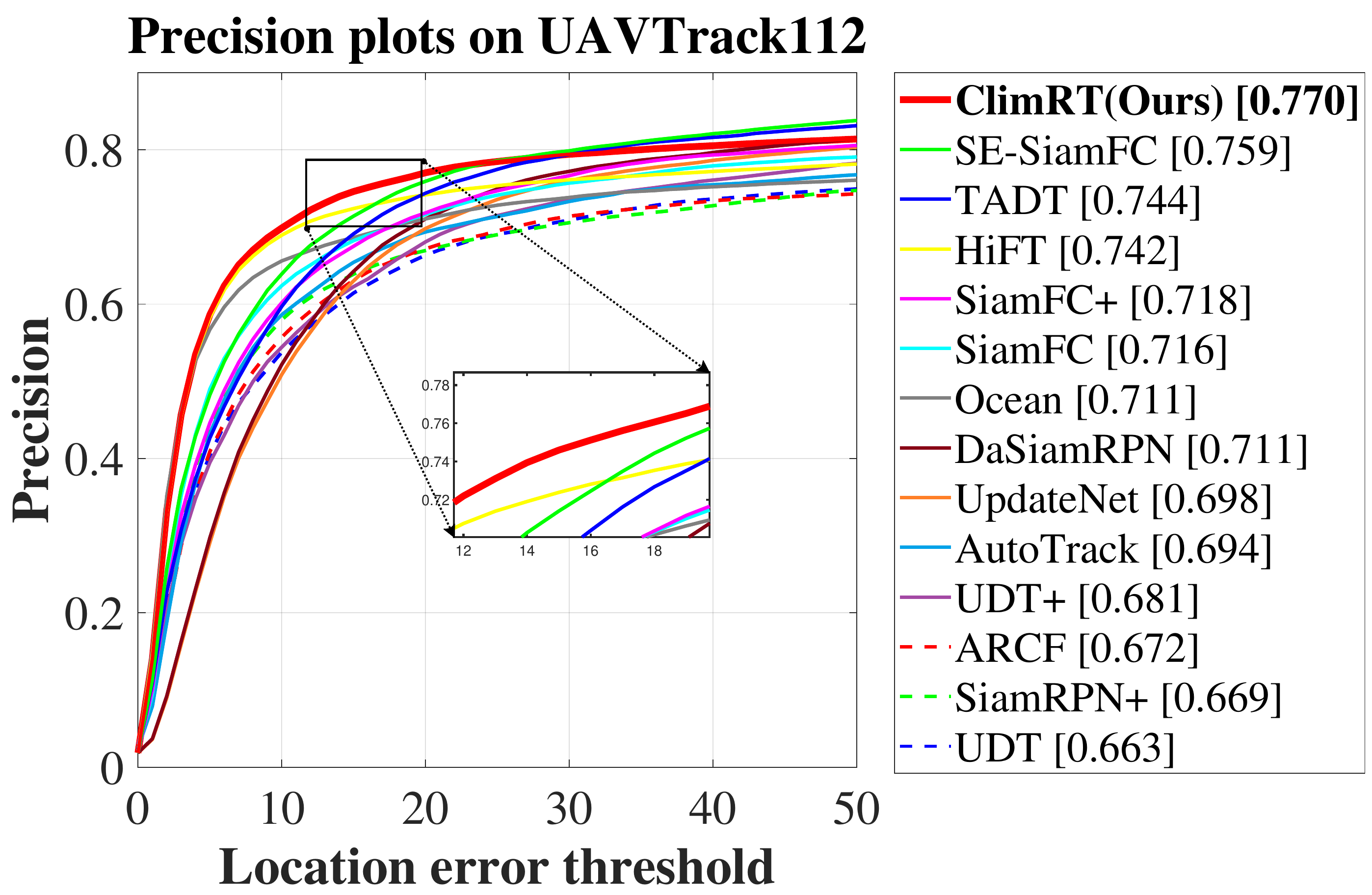}
}
\vfil
 \subfloat{
 \includegraphics[width=0.26\linewidth]{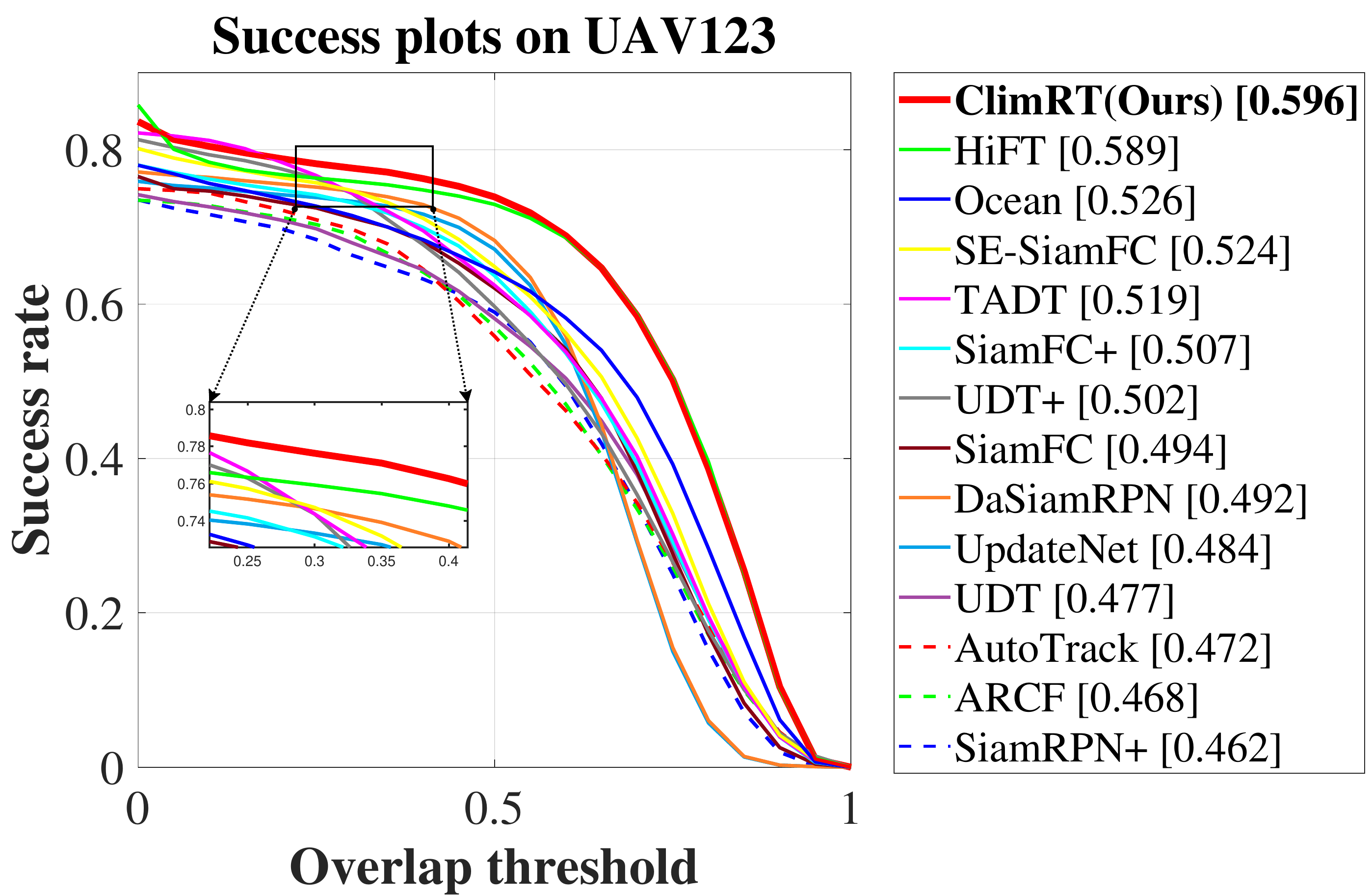}
 }
 \hfil
 \subfloat{
 \includegraphics[width=0.26\linewidth]{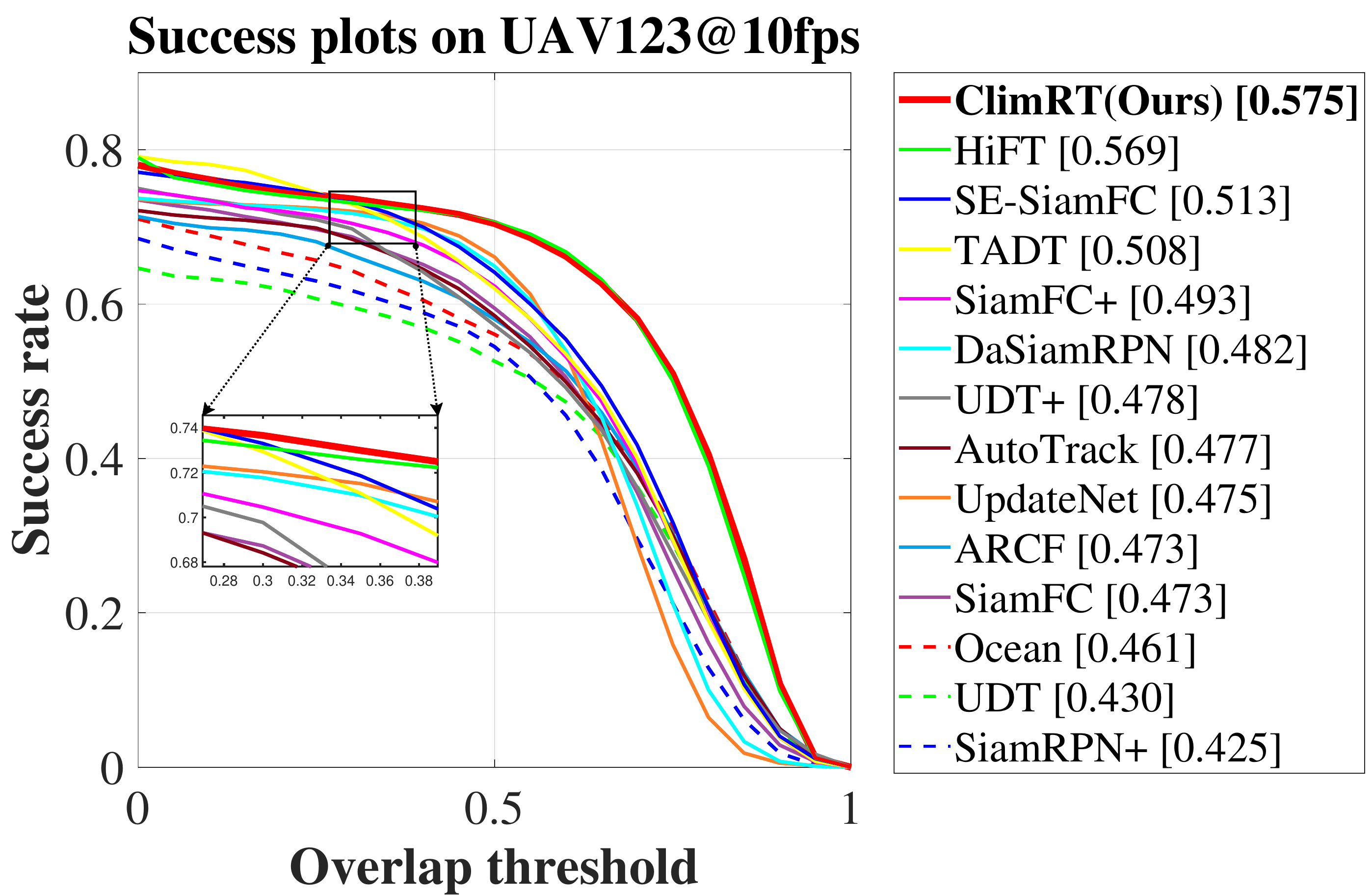}
 }
 \hfil
 \subfloat{
 \includegraphics[width=0.26\linewidth]{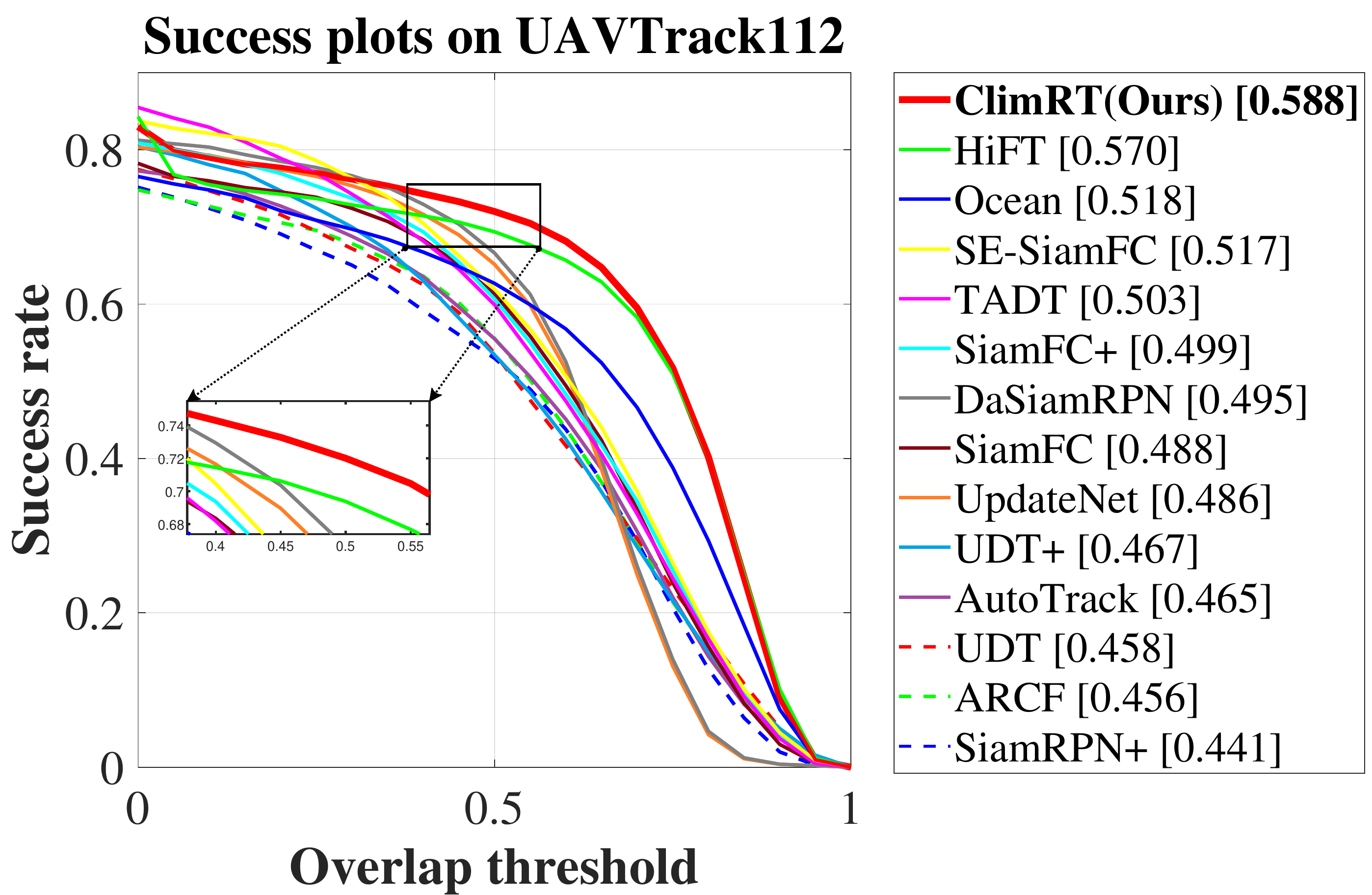}
}
 \caption
 {
  Overall performance of ClimRT and SOTA trackers on three well-known aerial tracking benchmarks. From left to right, the results on UAV123~\cite{MM2016UAV123}, UAV123@10fps~\cite{MM2016UAV123}, and UAVTrack112~\cite{Fu2022UAVTrack112} are presented. ClimRT achieves superior overall performance.
 }
 \label{5}
 \vspace{-10pt}
\end{figure*}

\subsubsection{UAV123}
UAV123~\cite{MM2016UAV123} is a large-scale UAV benchmark consisting of 123 high-quality sequences that contain a diverse range of challenging aerial scenarios. 
As illustrated in Fig.~\ref{5}, ClimRT outperforms other SOTA trackers in both precision (\textbf{0.798}) and success rate (\textbf{0.596}). 

\begin{table*}[hb]
 \vspace{-10pt}
\caption{Attribute-based evaluation of top 6 SOTA trackers. The best two performances are respectively highlighted by \textcolor{red}{red} and \textcolor{green}{green} color. ClimRT keeps achieving the best performance in typical UAV attributes.}
\centering
\renewcommand{\arraystretch}{0.95}
\newcommand{\tabincell}[2]{\begin{tabular}{@{}#1@{}}#2\end{tabular}}
\begin{tabular}{p{3.0cm} p{1.3cm}<{\centering} p{1.3cm}<{\centering} p{1.3cm}<{\centering} p{1.3cm}<{\centering }p{1.3cm}<{\centering} p{1.3cm}<{\centering} p{1.3cm}<{\centering} p{1.3cm}<{\centering}}
\toprule
\multirow{2}{*}{\diagbox{\textbf{Trackers}}{\textbf{Attributes}}} & \multicolumn{2}{c}{Aspect ratio change (ARC)} & \multicolumn{2}{c}{Full occlusion (FOC)} & \multicolumn{2}{c}{Partial occlusion (POC)} & \multicolumn{2}{c}{Out-of-view (OV)} \\
 & Prec. &  Succ. & Prec. &  Succ.& Prec. &  Succ. & Prec. &  Succ.\\
\midrule
SE-SiamFC~\cite{IS2021SESiam}  & 0.698 & 0.471 & 0.584 & 0.325 & 0.684 & 0.456 & 0.665 & 0.459 \\ 
SiamFC~\cite{LB2016SiamFC}  & 0.686 & 0.447 & 0.544 & 0.290 & 0.640 & 0.412 & 0.695 & 0.460 \\ 
Ocean~\cite{Zhang2020ocean}  & 0.668 & 0.471 & 0.597 & 0.355 & 0.638 & 0.443 & 0.637 & 0.441 \\
TADT~\cite{Li2019TADT}  & 0.666 & 0.455 & \textcolor{green}{0.608} & 0.338 & \textcolor{green}{0.693} & 0.472 & 0.625 & 0.444 \\ 
HiFT~\cite{Cao2021HiFT}  & \textcolor{green}{0.733} & \textcolor{green}{0.536} & 0.586 & \textcolor{green}{0.358} & 0.684 & \textcolor{green}{0.487} & \textcolor{green}{0.700} & \textcolor{green}{0.522} \\
\hline
\textbf{ClimRT (Ours)}   & \textcolor{red}{0.756} & \textcolor{red}{0.553} & \textcolor{red}{0.628} & \textcolor{red}{0.390} & \textcolor{red}{0.717} & \textcolor{red}{0.512} & \textcolor{red}{0.757} & \textcolor{red}{0.558} \\ 
\bottomrule
\end{tabular}
\label{table1}
 \vspace{-15pt} 
\end{table*}

\begin{figure}[tp]
	\centering
	\includegraphics[width=0.72\linewidth]{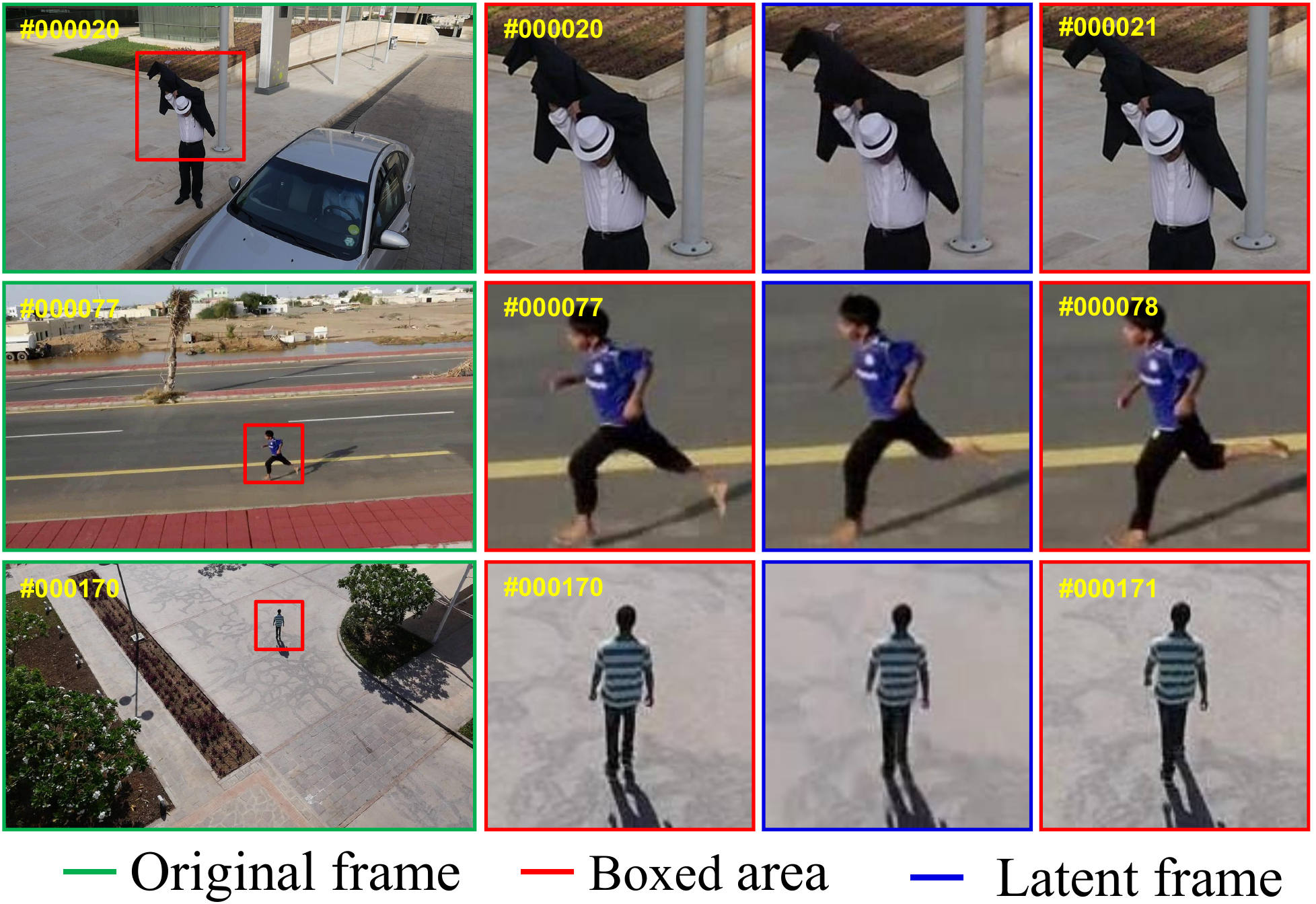}
	\caption{Continuity-aware latent interframe information mining results of three sequences from UAV@12310fps~\cite{MM2016UAV123}. The three rows correspond to \textit{person18}, \textit{person23}, and \textit{person12\underline{~}1}, respectively.
	\iffalse The satisfactory performance proves that ClimNet can effectively improve the consistency and continuity of information for objects at different scales.\fi}
	\label{6}
	\vspace{-22pt}
\end{figure}

\subsubsection{UAV123@10fps}
UAV123@10fps~\cite{MM2016UAV123} is created by downsampling from the original 30FPS recording. 
As a result, the continuity of information that can be extracted from the network is lower compared to UAV123. 
The results shown in Fig.~\ref{5} demonstrate that ClimRT can consistently achieve satisfactory performance, achieving the best precision (\textbf{0.760}) and success rate (\textbf{0.575}).

\subsubsection{UAVTrack112}
UAVTrack112~\cite{Fu2022UAVTrack112} is designed for aerial tracking and contains 112 real-world sequences with aerial-specific challenges.
As shown in Fig.~\ref{5}, ClimRT yields the best precision (\textbf{0.770}) and success rate (\textbf{0.588}).
The exceptional performance of ClimRT demonstrates that it is competent for real-world aerial tracking scenarios.

\vspace{-6pt}
\subsection{Continuity-Aware Latent Interframe Information Mining}
\vspace{-4pt}
The qualitative results of using ClimNet on three sequences from UAV123@10fps~\cite{MM2016UAV123} are presented in Fig.~\ref{6}.
The latent frames indicate the remarkable continuity-aware latent interframe information mining capability of ClimNet. 
Despite the low framerate limitation, ClimNet can still achieve ideal interframe information mining results, providing additional reliable information to the tracking network.

\begin{figure}[tbp]
	\centering
	\includegraphics[width=0.87\linewidth]{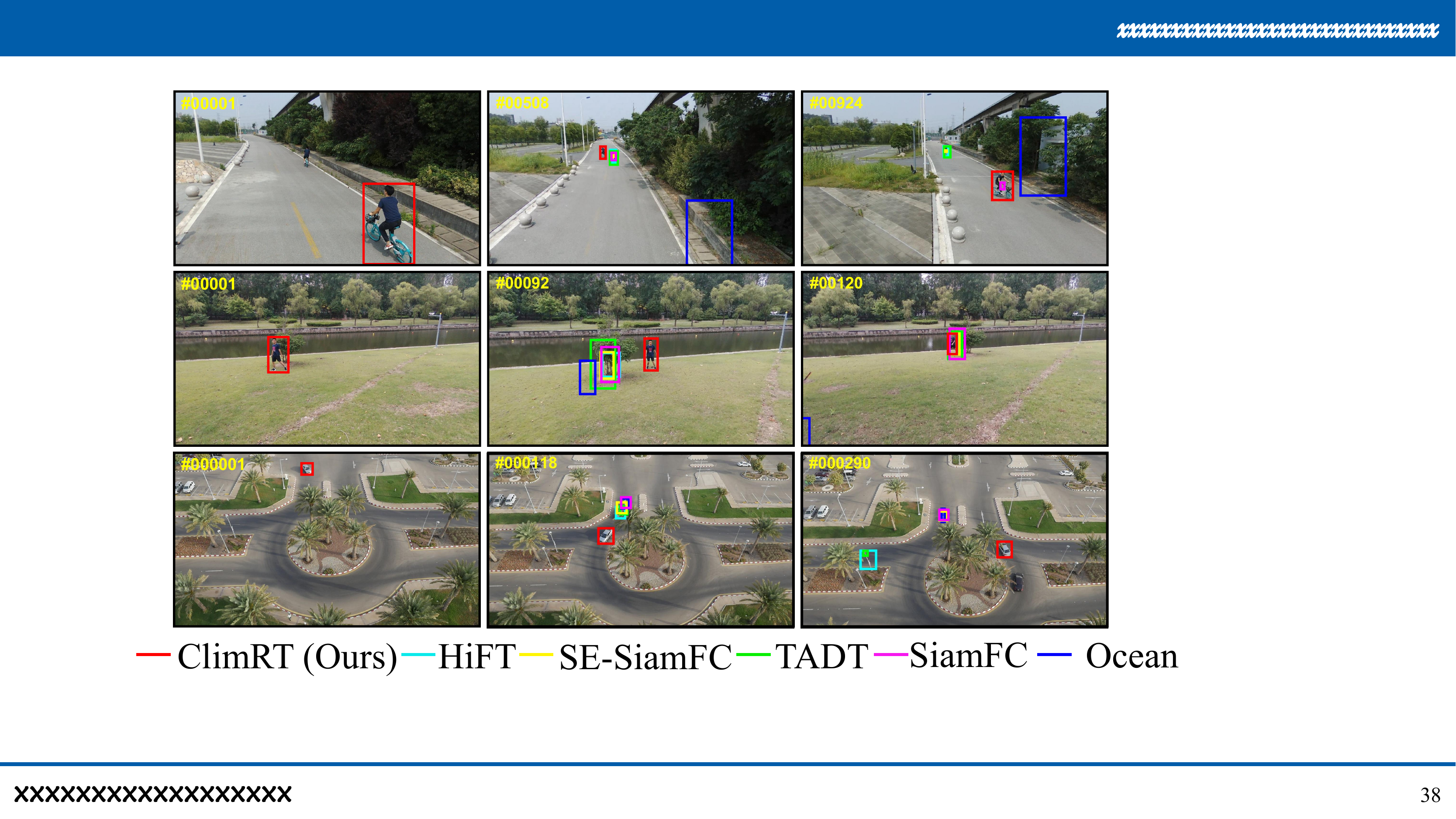}
	\caption{Qualitative comparison of the proposed ClimRT with the other top 5 trackers on three challenging sequences (\textit{bike6}, \textit{human5} from UAVTrack112~\cite{Fu2022UAVTrack112}, and \textit{car7} from UAV123@10fps~\cite{MM2016UAV123}). Owing to the reliable information produced by ClimNet, ClimRT can achieve robust performance under various challenges, especially in the presence of occlusion and aspect ratio change.}
	\label{7}
	\vspace{-20pt}
\end{figure}

\vspace{-6pt}
\subsection{Attribute-Based Comparison}
\vspace{-4pt}
To exhaustively evaluate ClimRT under various challenges, attribute-based comparisons are performed, as shown in TABLE~\ref{table1}.
ClimRT ranks first place in terms of both precision and success rate in comparison with other top 5 SOTA trackers. 
The satisfactory results demonstrate that the ClimNet can provide sufficient and reliable information to overcome mutation or even loss of object information. 
When the objects are occluded, ClimRT can learn more spatial-temporal information to discriminate the occluded objects. 
Furthermore,  ClimRT can effectively deal with the problem of perspective change in aerial scenarios due to the full mining of latent interframe information.
This is supported by its strong performance in aspect ratio change.
Generally, when encountering sudden changes in object state, ClimRT can effectively smooth the mutation process and supplement more information to enhance robustness.
As shown in Fig.~\ref{7}, ClimRT also achieves impressive performance in qualitative comparison with other SOTA trackers.

\begin{figure}[tbp]
	\centering
	\includegraphics[width=0.75\linewidth]{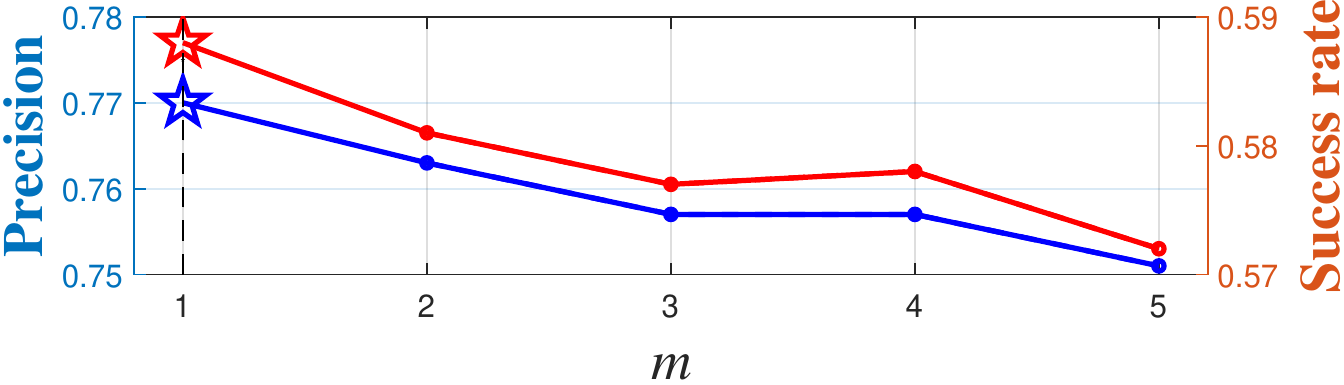}
	\caption{Parameter analysis of \textit{m} on UAVTrack112~\cite{Fu2022UAVTrack112}. When the \textit{m} = 1, ClimRT achieves the best overall performance.}
	\label{8}
	\vspace{-20pt}
\end{figure}

\vspace{-6pt}
\subsection{Ablation Study}
\vspace{-4pt}
To verify the effectiveness of each module of the proposed method, studies amongst ClimRT with different modules enabled are conducted on UAVTrack112~\cite{Fu2022UAVTrack112}. 
HiFT is considered as Baseline in this work.
For fairness, each variant of the tracker adopts the same training strategy and variables.

\subsubsection*{\bf Discussion on Continuity-Aware Latent Interframe Information Mining}
As shown in TABLE~\ref{table2}, the single LCT outperforms Baseline slightly. 
However, when GSTC and MS fusion blocks are integrated respectively, both precision and success rate are further improved compared to Baseline.
It proves that continuity-aware latent interframe information mining can efficiently enhance tracking robustness.
The adoption of ClimNet+LCT further results in the achievement of the best performance, exhibiting a \textbf{3.8\%} increase in precision and a \textbf{3.2\%} increase in success rate compared to Baseline.
These results all demonstrate the efficiencies of ClimNet and LCT in improving UAV tracking.
\subsubsection*{\bf Discussion on LCT}
As shown in TABLE~\ref{table2}, ClimNet+HiFT can improve the tracking performance by 1.1\% in precision.
However, when ClimNet is combined with LCT, the accuracy and robustness are significantly improved, up to \textbf{2.7\%} compared to ClimNet+HiFT.
This reflects the strong applicability of LCT to the entire network and its efficient information integration ability.

\begin{table}[b]
\vspace{-10pt}
 \centering
 \caption{Ablation study on various ClimRT components. $\Delta$ denotes the improvement compared with the Baseline.}
    \label{table2}
 \resizebox{\linewidth}{!}{
  \begin{tabular}{lcccc}
   \toprule
   Trackers & Prec.   & $\Delta_{Prec.}$(\%) & Succ. & $\Delta_{Succ.}$(\%) \\
  \midrule
   Baseline  & 0.742  & -  & 0.570  & -\\ 
   LCT & 0.747 & +0.7 & 0.573  & +0.5 \\
   GSTC+LCT & 0.754 & +1.6 & 0.575  & +0.9\\
   MS fusion+LCT & 0.759 & +2.3 & 0.581 & +1.9\\
   \hline
   ClimNet+HiFT & 0.750 & +1.1 & 0.577 & +1.2\\
  \hline 
     $\textbf{ClimNet}$ + $\textbf{LCT}$ ($\textbf{ClimRT}$) & $\textbf{0.770}$ & $\textbf{+3.8}$ & $\textbf{0.588}$ & $\textbf{+3.2}$ \\
   \bottomrule
      \end{tabular}
      }
\vspace{-10pt}
\end{table}

\vspace{-6pt}
\subsection{Parameter Analysis}
\vspace{-4pt}
The time spans between the previous and current search, \textit{i.e.}, \textit{m}, plays a crucial role in the reliability of ClimRT.
Setting a larger \textit{m} can obtain information over a longer time span, but it may also reduce the effectiveness of interframe information mining.
As illustrated in Fig.~\ref{8}, the value of \textit{m} is set from 1 to 5.
When \textit{m} is set to 1, the precision and success rate reach the highest score.
Besides, as \textit{m} continues to increase, the continuity between frames decreases. 
At the same time, the change of object state is intensified, which results in the lack of reliable object information, leading to a decline in performance.
Therefore, \textit{m} = 1 is deemed optimal for achieving the best performance.

\vspace{-6pt}
\subsection{Real-World Tests}
\begin{figure}[tbp]
	\centering
	\includegraphics[width=0.9\linewidth]{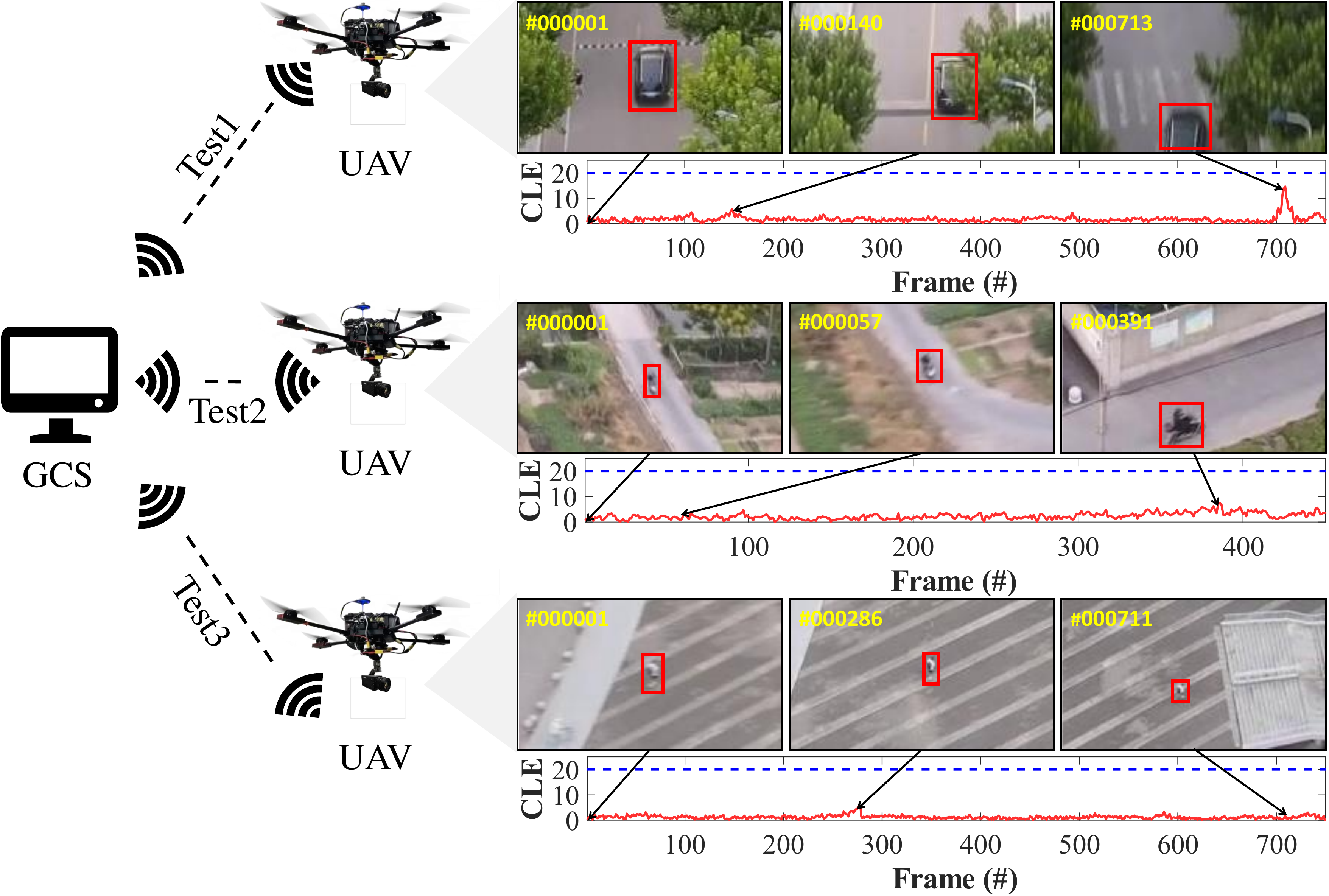}
	\caption{Visualization of real-world tests. The tracking results are marked with \textcolor{red}{red} boxes. The CLE score below the \textcolor{blue}{blue} dotted line is considered as reliable tracking in the real-world test.}
	\label{9}
	\vspace{-20pt}
\end{figure}

\vspace{-4pt}
The practicability of ClimRT is further testified in real-world applications. 
A ground control station (GCS) with an Intel \textit{i}9-9920X CPU and an NVIDIA TITAN RTX GPU receives and processes real-time images.
Then the processed tracking results are transmitted from the GCS to UAV for control via WiFi.
The proposed ClimRT achieves an average speed of 19.7 FPS during the test.
Fig.~\ref{9} presents three real-world aerial tests. 
The three tests contain several typical UAV tracking challenges, including occlusion and aspect ratio change.
It can be seen in Test 1 that when encountering occlusion, the tracking results exhibit some fluctuations, but are able to quickly restore stability.
In Test 2 and Test 3, ClimRT encounters object deformation caused by aspect ratio change.
Nevertheless, ClimRT can keep up with object even if its appearance changes suddenly.


\vspace{-2pt}
\section{Conclusions}
\vspace{-2pt}
In this work, a novel framework with continuity-aware latent interframe information mining is proposed to provide reliable and consecutive information for UAV tracking.
With the robust information mining and effective feature restoration capabilities of GSTC and MS fusion blocks, ClimNet can achieve satisfactory performance even in challenging aerial scenarios such as occlusion and aspect ratio change.
Besides, attributing to LCT, sufficient information can be efficiently integrated to improve information continuity.
Comprehensive experiments provide strong evidence that ClimRT can achieve robust performance in real-world and complex aerial tracking scenarios.
We are convinced that this work will promote the development of UAV tracking and foster the progress of autonomous UAV applications.

\vspace{-4pt}
\section*{Acknowledgment}
\vspace{-4pt}

This work is supported by the National Natural Science Foundation of China (No. 62173249), the Natural Science Foundation of Shanghai (No. 20ZR1460100).

\bibliographystyle{IEEEtran}

\bibliography{reference}


\end{document}